# The Deterministic Dendritic Cell Algorithm

Julie Greensmith[1] and Uwe Aickelin[1]

Intelligent Modeling and Analysis,
School of Computer Science,
University of Nottingham, UK, NG8 1BB.
`jqg, uxa@cs.nott.ac.uk`

**Abstract.** The Dendritic Cell Algorithm is an immune-inspired algorithm originally based on the function of natural dendritic cells. The original instantiation of the algorithm is a highly stochastic algorithm. While the performance of the algorithm is good when applied to large real-time datasets, it is difficult to analyse due to the number of random-based elements. In this paper a deterministic version of the algorithm is proposed, implemented and tested using a port scan dataset to provide a controllable system. This version consists of a controllable amount of parameters, which are experimented with in this paper. In addition the effects are examined of the use of time windows and variation on the number of cells, both which are shown to influence the algorithm. Finally a novel metric for the assessment of the algorithms output is introduced and proves to be a more sensitive metric than the metric used with the original Dendritic Cell Algorithm.

## 1 Introduction

Artificial Immune Systems (AISs) have developed significantly over the past five years, instigated by the creation of novel algorithms termed '2nd Generation AISs'. These AISs initially rely on interdisciplinary collaboration to use current research in immunology to produce algorithms which are both true to the underlying metaphor used as inspiration and perform well upon their resultant application domain. One such 2nd Generation AIS is the Dendritic Cell Algorithm (DCA), which is based on models of the dendritic cells (DCs) of the human immune system.

The original DCA was developed as part of the Danger Project [1], and formed the majority of Greensmith's thesis [3]. A prototype of the algorithm was first presented in 2005 [4] with a fully implemented real-time system version presented in 2006 [8]. The DCA has distinct advantages when applied to real-time computer security problems, as it has very low CPU processing requirements and does not require extensive training periods. All versions of the DCA to date have used a relatively large number of parameters and stochastic elements, such as random selection of cells and variable thresholds. Setting these parameters to the appropriate values has always been somewhat arbitrary, and thus has left the algorithm open to various criticisms. The use of various probabilistic elements was in part an artifact of the use of the Twycross' `libtissue` framework for the initial algorithm development. While this framework is useful for the rapid development of such AISs, one of the drawbacks for the DCA is the sheer amount of interacting entities.

As a result, it is still unclear which parts of the algorithm are responsible for its performance and for its time-dependent correlation properties. In order to push forward the DCA as a serious contender within biologically inspired computation, a thorough analysis of the algorithm itself must be performed: a task too complex when implemented within a large framework. Insight is needed into exactly what each component of the algorithm does and how detection is actually achieved. Despite avoiding a theoretical approach so far, the time has come to pick apart this algorithm and to break it down into a controllable deterministic system which is more accessible for the performance of various computational analyses and the various parameter relationships explored.

The aim of this paper is to describe, implement, and test a deterministic DCA (dDCA) to uncover its inner relationships and function. This paper is structured as follows, with background information present in section 2, section 3 describing the dDCA and the new metric $K_\alpha$. Experiments are described in section 4, with a discussion of results and conclusions presented in sections 5 and 6 respectively.

## 2  DCA Overview

Metaphorically, DCs are the crime-scene investigators of the human immune system, traversing the tissue for evidence of damage - namely signals, and for potential suspects responsible for the damage, namely antigen. More information regarding the function of natural DCs can be found in [10] with a distilled version for computer scientists presented in [3]. The DCA is derived from an abstract model of DC biology resulting in a population based algorithm which provides robust detection and correlation. Different cells process signals acquired over different time periods, generating individual 'snapshots' of input information which are correlated with antigens. The original DCA is described in detail in numerous sources including [7] and [3].

The majority of research performed with the DCA has been within the sphere of security. In particular, the works of Greensmith et al. have focussed on computer security applications. The algorithm to date has been successfully applied to port-scan detection [8] [6] [5], and upon comparison to a self organizing map performed well on the large dataset used, classifying 13 million antigens in under 100 seconds. In addition to her work, the DCA has also been applied to the detection of a novel threat on the internet, botnets [2], where the DCA produced high rates of true positives and low rates of false positives in comparison to a statistical technique. Outside of computer security Kim et al. have successfully applied the DCA to the detection of misbehaviour in wireless sensor networks, where again the algorithm showed a lot of promise. More recently in the work of Lay and Bate [9], the DCA is applied to the detection of overruns in the scheduling of processes, again with success.

The DCA is also showing promise in the area of robotic security as demonstrated by Oates et al. [11]. A proof of concept experiment is performed to demonstrate that the DCA could be used for basic object discrimination in a controlled environment. The same researchers have now extended this research into the theoretical domain [12] through frequency tuning analysis. This research has highlighted that the DCA exhibits filter properties and also suggests the importance of the lifespan limit. Their research also contains two optimizations of the DCA which are used in this paper, namely a real valued representation of individual DC output and tissue centric processing of signals.

## 3 The Deterministic DCA (dDCA)

In this section the dDCA is formally described followed by a discussion of the modified features. In order to produce the deterministic version, it is necessary to make a number of assumptions and modifications to the original DCA:

- Both signals and antigen are required for the system to correctly function. If no signals are used, then the DCs will not exceed their lifespan limit and will not be able to present antigen. If no antigen are used, then the context has no subject.
- At minimum two signal categories are required, an activating signal and an inhibitory signal - the danger and safe signal respectively.
- A uniform distribution of lifespan values is used across the population. This allows for the study of the time-window effect in a repeatable and controllable manner.
- To provide reproducibility and for the ease of sensitivity a reduction in parameters is required from those used with the original DCA. As a result explicit antigen storage and sampling of the antigen population is removed, with all antigen data sampled by the DC population.
- Each DC in the population is exposed to identical input signal data and would process these signals in an identical manner. This results in the optimisation of the signal processing procedure, as the output signal values are calculated only once for the entire population, as suggested by Oates et al. [12].
- The output context value of an individual DC is reduced to one factor, $\bar{k}$, which negative numbers indicate a safe context and positive numbers indicating analogous to the previously used mature context. This is also derived from the theoretical analysis provided in Oates et al. [12].

One further modification is proposed for use with this system. This is the incorporation of an antigen profile. In previous implementations of the DCA, the string type antigens are stored in an 'antigen vector' data structure. This required the random selection of antigen by each DC and antigen overwriting. To ensure exact reproducibility the random sampling and storage is replaced by a simple array. In this array the value of the antigen is stored with the number of times a DC has collected antigens of this type. This reduces the required overhead as no dynamic memory management is required and leaves no concerns over denial of service due to the potential threat of antigen flooding.

Previous versions of the DCA featured in excess of 10 parameters, each of which were derived from empirical biological observation and through sensitivity analysis. The resultant algorithm contains three parameters. Firstly, the number of DCs must be defined - this is set to 100 as previously, but is experimented with in Section 4. Secondly the weighting schema for the signal processing. The signal processing equation used previously is modified for use with simplified weight values. As with the original DCA, the input signals are transformed to output signals. However a different procedure is needed as the processing is performed in the tissue, the incorporation of k reduces the outputs from three to two and this is coupled with the reduction to two signal categories. The new signal processing procedure is shown in Equations 1 and 4, where S and D is the input value for the safe and danger signals respectively with 2 and 3 showing subsequent derivation thereof, c is the interim costimulation output signal and k is the interim context output value. Pseudocode for the implemented dDCA is given in Algorithm 1.

$$csm = S + D \tag{1}$$
$$k = (mature - semi) \tag{2}$$
$$k = (D - S) - S \tag{3}$$
$$k = D - 2S \tag{4}$$

```
input  : Antigen and Signals
output: Antigen Types and cumulative k values
set number of cells;
initialise DCs();
while data do
    switch input do
        case antigen
            antigenCounter++;
            cell index = antigen counter modulus number of cells ;
            DC of cell index assigned antigen;
            update DC's antigen profile;
        end
        case signals
            calculate csm and k;
            for all DCs do
                DC.lifespan -= csm;
                DC.k += k;
                if DC.lifespan <= 0 then
                    log DC.k, number of antigen and cell iterations ;
                    reset DC();
                end
            end
        end
    end
end
for each antigen Type do
    calculate anomaly metrics;
end
```
Algorithm 1: Pseudocode of the deterministic DCA.

## 3.1 Anomaly Metrics: MCAV and $K_\alpha$

The mature context antigen value (MCAV) is calculated once all data is processed, derived from the output of the cells collected during run-time. This value is generated for each antigen type ($\alpha$), where $\alpha$ is defined as a set of antigens of identical value. As the name suggests, the MCAV is a measure of the proportion of antigen presented by a fully mature cell as shown in Equation 5, where $MCAV_\alpha$ is the MCAV for antigen type $\alpha$, M is the number of 'mature' antigen of type $\alpha$, and Ag is the total amount of antigen presented for antigen type $\alpha$.

$$MCAV_\alpha = \frac{M}{Ag} \quad (5)$$

This metric returns a value between zero and one, where the probability of an antigen type being anomalous increases as this value tends to one. This is a convenient, normalised output, to which an anomaly threshold can be applied. However, it fails to encapsulate the magnitude of the difference between positive and negative values of the presented k. In the MCAV calculation a value of k of -1 is treated in exactly the same manner as a value of -200. The algorithm provides this information, hence it may be fruitful to incorporate this information into a more sophisticated metric.

$\overline{K_\alpha}$ is implemented with the dDCA, and uses the magnitudes of the $\overline{k}$ values. This generates real valued anomaly scores and may assist in the polarisation of normal and anomalous processes. The process of calculating this anomaly score is shown in Equation 6, where $k_m$ is the $\overline{k}$ value for $DC_m$, $\alpha_m$ is the number of antigen presented of type $\alpha$ by $DC_m$.

$$\overline{K_\alpha} = \frac{\sum_m k_m}{\sum_m \alpha_m} \quad (6)$$

As this equation returns real valued numbers dependent on the actual values of the input signals used, we propose a method for defining an anomaly threshold, to allow for the classification of the antigen types analysed. This can be performed if the signals are known a priori. The number of signal instances and the equivalent processed total sum of the input signals. The threshold, $T_K$, is defined in Equation 7 with $S_K$, the weighted sum of all input signals, defined in Equation 8, where $I_s$ is the number of pairs of signal instances, $\bar{i}$ is the mean number of iterations per cell incarnation, and D and S representing danger and safe signal values.

$$T_K = \frac{S_K}{I_s} * \bar{i} \quad (7)$$

$$S_K = \sum_{I_s} D - 2 \sum_{I_s} S \quad (8)$$

Once $T_K$ is applied to the $\overline{K_\alpha}$ values, antigen types with a value of over this thresh- old are classed as anomalous, and lower values classed as normal. If required, true and false positives can be derived from this information. A similar threshold can be derived from the MCAV, using the ratio of total danger signals to total safe signals present in the used dataset.

## 4 Experimental Analysis

### 4.1 Introduction

In this section initial tests are performed using the dDCA. This involves re-visiting a past dataset, namely the ping scan data used in Greensmith et al. [7] with one randomly selected set used to test the algorithm. In these experiments two aspects of the algorithm's function are examined:

- E0: A validation exercise to ensure the dDCA is correct.
- E1: The influence of variation in the number of cells.
- E2: Examination of 'time windows' and their effects on performance.

### 4.2 Testing Dataset

For these experiments one safe and one danger signal are used to provide the context information. As opposed to contriving artificial data, a dataset containing an outbound port scan is used. The object of using this data is that it is real-world data yet it is also relatively small, with approximately 25,000 antigens and 38 sets of danger and safe signal instances. The data is derived from a monitored remote shell session, where antigens are derived from process ID numbers and signals from monitored attributes of machine behaviour. Specifically, the danger signal is the rate of sending of outbound network packets, with the safe signal being the inverse rate of change of the packet sending rate. For more information of the necessity of these signals for port scan detection and for the mechanisms involve in port scanning please refer to [3].

In this dataset signals are updated once per second, with antigens generated as processes produce system calls. Both signals are normalised within a range of 0 to 50, based on maximum values derived in preliminary experiments. A graph of these signals is shown in Figure 1(a), where the mean danger signal value is 15.0 and mean safe signal value is 21.8. In terms of antigens, four processes of interest are captured by the antigen generator. These processes include two anomalous processes namely nmap the port scan process and pts a parent process of the nmap. Also included are two normal processes including sshd the remote shell facilitator process and bash the process of the actual monitored remote shell. The aim of the dDCA for these experiments is to produce high MCAV and $K_\alpha$ for the nmap and pts with lower values for the bash and sshd processes.

### 4.3 Experimental Setup

The deterministic DCA has two parameter values namely the number of cells and the lifespan limit. Unless specified otherwise, all experiments described use 100 artificial DCs with a maximum lifespan limit of 100 csm signal units. The increments of the lifespans are derived from the maximum limit divided by the number of cells. This is used to ensure an equivalent range of cells are present in each experiment. The $T_K$ value used for this experiment is calculated as shown in Equation 9 , where the number of signal instances is 38 and the mean number of iterations per cell incarnation is 2.

Table 1. MCAVs produced for dDCA versus Original DCA (mean of 3 runs).

| Process ID | Original DCA | dDCA |
|---|---|---|
| nmap | 0.999 | 0.969 |
| pts | 0.901 | 0.830 |
| bash | 0.711 | 0.623 |
| sshd | 0.070 | 0.202 |

The anomaly threshold for the MCAV is set to 0.69 based on the ratio of danger to safe signals within the dataset. The signal processing schema used is the one described previously in Equation 4 For the implementation, the dDCA is coded in C (gcc 4.0.1), with all experiments run on a 2.2 GHz MacBook Intel Core 2 Duo.

$$-57.4 = \frac{-1090}{38} * 2 \tag{9}$$

4.4 E0: Validation

Before the dDCA can be used for these experiments, it must first be validated against the results generated by the original DCA. For this purpose, the results presented for the original DCA are derived from data used for Chapter 6 of [3]. The results of one run of the dDCA with default parameters are compared with three runs of the original DCA, with the MCAV results generated presented in Table 1. As shown in this table, the same trends are evident in both datasets. However, less polarisation between the normal and anomalous processes is shown with the dDCA. Despite such discrepancies, as similar trends are shown, we are confident that the dDCA is valid as a form of DCA.

4.5 E1: Cell Number Experiments

In this series, the number of cells used to process data are varied between runs. The set of cell numbers used is n = $\{1, 5, 10, 50, 100, 500, 1000, 5000\}$. Based on past sensitivity analyses of the cell numbers we expect the greatest variation between 1 and 100 cells. In addition to exploring this relationship, this experiment is used to generate statistics regarding the mean behaviour of the cell population. During these experiments, the number of antigen presented per cell per iteration, the number of iterations per lifespan and the number of cell resets are collated and mean values are calculated. Additionally, these experiments are timed to gain some insight into the scalability of the algorithm.

Both the MCAV and $K_\alpha$ values are shown for the four processes of interest for each cell number and we can use this information to assess the differences between the two output metrics. We predict that the real valued magnitude of $K_\alpha$ will produce more polarised results as it will provide discrimination between borderline cases and the more extreme, which of course is merely represented as 0 or 1 for the MCAV.

The results for the cell number experiments are shown in Figures 1(b) and 1(c). A graph of the timing results for the experiments are presented in Figure 1(d). Statistics regarding the cell behaviour information are presented in Table 2.

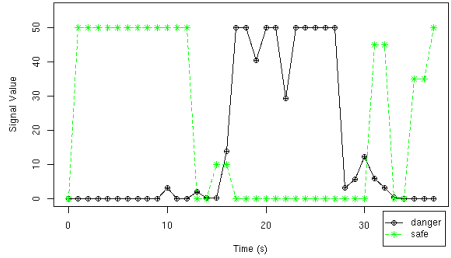
(a) Input Signals for the 38s session

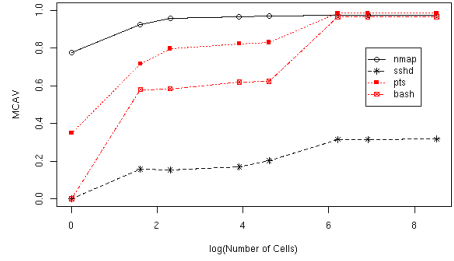
(b) MCAV of varying cell numbers

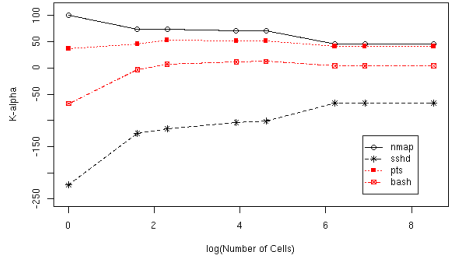
(c) $K_\alpha$ of varying cell numbers

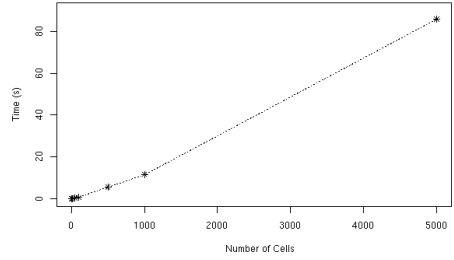
(d) Execution Times of varying cell numbers

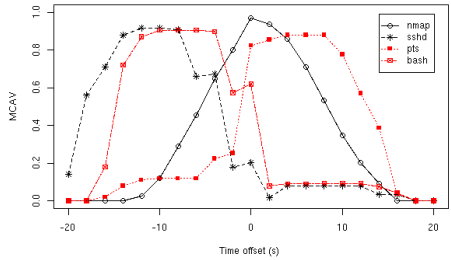
(e) MCAV for time-shifts

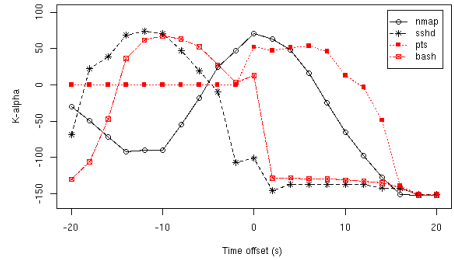
(f) $K_\alpha$ for time-shifts

Fig. 1. The input signal data is displayed in (a) with results for both series of experiments given in (b) to (f). Figures (b) and (c) show the MCAV and $K_\alpha$ values across a range of cell numbers plotted on a log-scale, (d) shows the execution times for varying the cell numbers, with (e) and (f) showing the MCAV and $K_\alpha$ with varying time delays.

Table 2. Cell behaviour statistics.

| Cell Number | Mean iterations | Mean incarnations |
|---|---|---|
| 1 | 3.7 | 19.0 |
| 5 | 2.3 | 10.0 |
| 10 | 2.1 | 8.7 |
| 50 | 1.9 | 9.9 |
| 100 | 1.8 | 10.1 |
| 500 | 1.1 | 17.4 |
| 1000 | 1.1 | 17.4 |
| 5000 | 1.0 | 17.5 |

### 4.6  E2: Time Window Experiments

It is assumed that the DCA performs correlation between antigen and signals based on time windows. These experiments are designed to ascertain if this is indeed the case. The nature of the time window effect created by the population of DCs is examined by shifting the position of the signals within the dataset. Each cell in the population has a lifespan, which defines the quantity of input signals the cell can process per incarnation. Having variable time windows should add robustness when the signals occur after the antigens, but we expect a reduction in DCA performance should the signals occur before the antigen.

While the cells create a type of moving average for the signals, this does not extend before the cell is initialised, and therefore signals appearing before antigen may result in a poor performance. A total of 20 extra datasets are created, with a maximum shift of 20 second for the signals before and after the original position, at two second increments. As with E1, both the MCAV and $K_\alpha$ values are calculated for each process of interest. These results are presented in Figures 1(e) and 1(f).

## 5  Experimental Analysis

### 5.1  E1: Cell Numbers

In E0 the dDCA is validated as fit for purpose. Subsequently when the number of cells is varied in E1 a noticeable effect on the performance of the DCA is indicated as shown in Figures 1(b) and 1(c). When the MCAV is used as the anomaly metric, an increase in the number of cells causes an increase in the MCAV for both pts and bash, though sshd and nmap do not increase to the same magnitude. The same trends are evident though less noticeable when using $K_\alpha$ for the bash and pts processes. This may be because it is difficult to assess if these processes, the parent processes of the nmap scan process, are actually anomalous or normal given that they have involvement in facilitating the scan itself. These two processes are borderline cases, and it appears that $K_\alpha$ provides improved information for this type of input data.

The sshd process which does not assist the scan has consistently low $K_\alpha$ values, well below the derived threshold of -57.4. It is interesting to note that as the number of cells used increases, the resultant output values converge. One possible explanation for this is that the lifespan limit is set incorrectly and maybe an improvement could be made if the range of these thresholds also increase in proportion to the number of cells.

Another explanation is that once the number of cells exceeds a certain limit, the capacity of the system exceeds the requirements of the input data, and therefore no matter how many extra cells are added, the resultant values remain similar. This is also shown in the summary statistics of the cell behaviour presented in Table 1. The results of the timed experiments are also encouraging, giving that the relationship between the number of cells and the execution time appears to be linear.

### 5.2 E2: Time Windows

The results of experiment E2 also show similar trends in comparison between MCAV and $K_\alpha$, with the $K_\alpha$ values representing more precisely the classification of these processes. Therefore, $K_\alpha$ will be used in future for the assessment of our DCA experiments both empirical and theoretical. A marked difference is shown in particular for the nmap process between time offset -20 and zero and for the pts process also between -20 and zero.

Examination of the pts graphs show a moderately low MCAV value, yet when $K_\alpha$ is used, this value looks to remain stable at a level of 0. This could indicate that the pts process exhibits minor fluctuations around this point, with these fluctuations amplified by the binary classification of cells used in the MCAV, with $K_\alpha$ showing to be more sensitive to encapsulating such fluctuations.

In terms of the time window analysis two conclusions can be drawn from these graphs. Firstly, when the signals are delayed (time offset of 0 to 20), correct classification continues for almost 10 seconds, until the anomalous processes are classified as normal as they fall below $T_K$. Interestingly, improved results are shown with a delay of 2-4 seconds - which is equivalent to the average number of cell iterations per lifespan. Potentially the range of acceptable delay may be linked to a relationship between the number of iterations and the lifespan range itself, to which a formal analysis may be able to prove. Within the applications of the DCA in security so far, the signals are always updated after the antigens are generated, indicating one reason for why the DCA functions in the manner shown previously. These results suggest that the dDCA has the potential to be error tolerant to at least a five second lag in signal data, which is a desirable property for any behaviour based anomaly detection approach, as this reflects the situation often seen in real world intrusion data.

The opposite effect is shown when the signals are advanced ahead of the antigens. For the MCAV results both sets of processes, normal and anomalous, are classified incorrectly between time offset -20 and 0. A similar effect is seen for $K_\alpha$ for the same offset values. One explanation for this effect is that whilst cells produce a type of moving average, this is derived from information in only one direction i.e. the cells cannot incorporate information received before the start of their current incarnation. Therefore a reincarnated cell can only have knowledge of the signals which occur after its generation. While these results are interesting, a more formal analysis with contrived and controllable data must be performed in future in order to corroborate this tenet. This mirrors what is shown with natural DCs, as pathogenic infection (i.e. the presence of antigen) always occurs before the generation of danger signals.

## 6 Conclusions

In this paper a deterministic version of the DCA is proposed, implemented and tested. In addition to changes in the algorithm a new metric for the system's evaluation is proposed namely $K_\alpha$ which takes into account the magnitude of the output values produced by the DC population. The dDCA is compared to the original DCA using a port scan dataset used previously with the DCA. We are satisfied that while that results are not identical the values show similar trends, indicating that the essence of the DCA is housed within the deterministic version. This version has several advantages, including the ability to replay experiments exactly, predictability of output and the reduction in the number of parameters required. All such factors have resulted in a version of the DCA which is simple to implement and can produce reliable, consistent results.

One of the remaining parameters of the dDCA is the number of cells used. As this number increases, discrimination between the processes is less obvious. While the cause of this effect still remains unclear it has given us insight into the limits of the system as it appears that there is a saturation point. For this particular dataset, this point is at 500 cells shown for both the MCAV and $K_\alpha$. The metric $K_\alpha$ is tested for the first time in this experiment and is shown to be more sensitive to the minor fluctuations in the resulting output of the cells and provides a more precise overview of the classification of the various antigen types. To assess the implications of $K_\alpha$, this metric should be applied to a wider range of problems.

Finally, timing discrepancies between signals and antigen are performed. As a result it is shown that should there be a delay for the input signals, within a tolerance range the dDCA can cope well with this delay. A potential relationship between the lifespan maximum limit and the number of iterations per cell incarnation may exist, though a more formal analysis is required to verify this effect. Conversely, if the signal data is advanced, severe misclassifications can occur, hence suggesting that the dDCA should not be applied to data where there is the potential for delayed antigen as performance may be impaired.

As future work we intend to further explore this new instantiation of the DCA. This investigation will involve a more in-depth study of the inherent relationships present within the algorithm in addition to extensive testing both on a range of real-world and synthetic data, and in comparison with other standard techniques such as support vector machines. This has the aim of selecting such parameters appropriately no matter what the application. In conclusion, the dDCA is a comparable and controllable form of the DCA and is a powerful tool necessary to further the understanding of this interesting immune-inspired algorithm.

## Acknowledgements

This research is supported by the EPSRC (EP/D071976/1). Code optimisations courtesy of Gianni Tedesco.

# References

1. U. Aickelin, P. Bentley, S. Cayzer, J. Kim, and J. McLeod. Danger theory: The link between AIS and IDS. In Proc. of the 2nd International Conference on Artificial Immune Systems (ICARIS), LNCS 2787, pages 147–155. Springer-Verlag, 2003.
2. Y. Al-Hammadi, U. Aickelin, and J. Greensmith. DCA for detecting bots. In to appear in Proc. of the Congress on Evolutionary Computation (CEC), page tba, 2008.
3. J. Greensmith. The Dendritic Cell Algorithm. PhD thesis, School of Computer Science, University Of Nottingham, 2007.
4. J. Greensmith, U. Aickelin, and S. Cayzer. Introducing Dendritic Cells as a novel immune-inspired algorithm for anomaly detection. In Proc. of the 4th International Conference on Artificial Immune Systems (ICARIS), LNCS 3627, pages 153–167. Springer-Verlag, 2005.
5. J. Greensmith, U. Aickelin, and J. Feyereisl. The DCA-SOMe comparison: A comparative study between two biologically-inspired algorithms. Evolutionary Intelligence: Special Issue on Artificial Immune Systems, accepted for publication, 2008.
6. J. Greensmith, U. Aickelin, and G. Tedesco. Information fusion for anomaly detection with the DCA. Information Fusion, in print, 2008.
7. J. Greensmith, U. Aickelin, and J. Twycross. Articulation and clarification of the Dendritic Cell Algorithm. In Proc. of the 5th International Conference on Artificial Immune Systems (ICARIS), LNCS 4163, pages 404–417, 2006.
8. J. Greensmith, J. Twycross, and U. Aickelin. Dendritic cells for anomaly detection. In Proc. of the Congress on Evolutionary Computation (CEC), pages 664–671, 2006.
9. N. Lay and I. Bate. Improving the reliability of real-time embedded systems using innate immune techniques. Evolutionary Intelligence: Special Issue on Artificial Immune Systems, 2008.
10. M. Lutz and G. Schuler. Immature, semi-mature and fully mature dendritic cells: which signals induce tolerance or immunity? Trends in Immunology, 23(9):991–1045, 2002.
11. R. Oates, J. Greensmith, U. Aickelin, J. Garibaldi, and G. Kendall. The application of a dendritic cell algorithm to a robotic classifier. In Proc. of the 6th International Conference on Artificial Immune Systems (ICARIS), LNCS 4628, pages 204–215, 2007.
12. R. Oates, G. Kendall, and J. Garibaldi and. Frequency analysis for dendritic cell population tuning: Decimating the dendritic cell. Evolutionary Intelligence: Special Issue on Artificial Immune Systems, 2008.